\documentclass[conference]{IEEEtran}
\IEEEoverridecommandlockouts
\usepackage{cite}
\usepackage{amsmath,amssymb,amsfonts}
\usepackage{algorithmic}
\usepackage{graphicx}
\usepackage{textcomp}
\usepackage{xcolor}
\usepackage{hyperref}
\usepackage{graphicx}
\usepackage{multirow}
\usepackage{booktabs}
\usepackage{multirow}
\usepackage{bbding}
\usepackage{pifont}
\usepackage{wasysym}
\usepackage{amssymb}
\usepackage{footnote}
\usepackage{subfig}
\def\BibTeX{{\rm B\kern-.05em{\sc i\kern-.025em b}\kern-.08em
    T\kern-.1667em\lower.7ex\hbox{E}\kern-.125emX}}
\begin{document}
 \makeatletter
    \newcommand{\linebreakand}{%
      \end{@IEEEauthorhalign}
      \hfill\mbox{}\par
      \mbox{}\hfill\begin{@IEEEauthorhalign}
    }
    \makeatother

\title{LAM3D: Leveraging Attention for Monocular 3D Object Detection\\
}

\author{\IEEEauthorblockN{Diana-Alexandra Sas}
\IEEEauthorblockA{\textit{Technical University of Cluj-Napoca} \\
\textit{Faculty of Automation and Computer Science}\\
Cluj-Napoca, Romania \\
sas.cr.diana@student.utcluj.ro}
\and
\IEEEauthorblockN{Leandro Di Bella}
\IEEEauthorblockA{\textit{Vrije Universiteit Brussel} \\
\textit{ETRO}\\
Brussels, Belgium \\
leandro.di.bella@vub.be}
\and
\IEEEauthorblockN{Yangxintong Lyu}
\IEEEauthorblockA{\textit{Vrije Universiteit Brussel} \\
\textit{ETRO}\\
Brussels, Belgium \\
yangxintong.lyu@vub.be}
\linebreakand % <------------- \and with a line-break
\IEEEauthorblockN{Florin Oniga}
\IEEEauthorblockA{\textit{Technical University of Cluj-Napoca} \\
\textit{Computer Science Department}\\
Cluj-Napoca, Romania \\
florin.oniga@cs.utcluj.ro}
\and
\IEEEauthorblockN{Adrian Munteanu}
\IEEEauthorblockA{\textit{Vrije Universiteit Brussel} \\
\textit{ETRO}\\
Brussels, Belgium \\
adrian.munteanu@vub.be}
}
%\and
% \and

\maketitle

\begin{abstract}
% Since the introduction of the self-attention mechanism and the adoption of the Transformer architecture for Computer Vision tasks, the Vision Transformer-based architectures gained a lot of popularity in the field, being used for tasks such as image classification, object detection and image segmentation. Nevertheless, the existing solutions for the Monocular 3D Object Detection vision task aren't leveraging the power of the attention mechanism, using Convolutional Neural Networks (CNNs) as a feature extraction backbone instead. In this paper, we present {LAM3D}, a framework that \underline{L}everages self-\underline{A}ttention mechanism for \underline{M}onocular \underline{3}D object \underline{D}etection. To do so, the proposed method is built upon a Pyramid Vision Transformer v2 (PVTv2) as feature extraction backbone. The training and evaluation of the proposed method is done on KITTI 3D Object Detection Benchmark, proving the applicability of the proposed solution in the autonomous driving domain and outperforming reference methods. An ablation study and extensive experiments evaluate the impact brought by the attention mechanism for this task and demonstrate systematic improvements over the equivalent architecture.
Since the introduction of the self-attention mechanism and the adoption of the Transformer architecture for Computer Vision tasks, the Vision Transformer-based architectures gained a lot of popularity in the field, being used for tasks such as image classification, object detection and image segmentation. 
However, efficiently leveraging the attention mechanism in vision transformers for the Monocular 3D Object Detection task remains an open question.
In this paper, we present {LAM3D}, a framework that \underline{L}everages self-\underline{A}ttention mechanism for \underline{M}onocular \underline{3}D object \underline{D}etection. To do so, the proposed method is built upon a Pyramid Vision Transformer v2 (PVTv2) as feature extraction backbone and 2D/3D detection machinery. 
We evaluate the proposed method on the KITTI 3D Object Detection Benchmark, proving the applicability of the proposed solution in the autonomous driving domain and outperforming reference methods.
Moreover, due to the usage of self-attention, LAM3D is able to systematically outperform the equivalent architecture that does not employ self-attention.
\end{abstract}

% \begin{IEEEkeywords}
% component, formatting, style, styling, insert
% \end{IEEEkeywords}

\section{Introduction}
Object Detection is one of the fundamental tasks in Computer Vision, consisting of detecting and locating objects of interest of a specific class within an image or video. The 2D detection of an object means determining the object's position within the image in terms of a 2D bounding box and classifying the object in a specific category. The state-of-the-art approaches for 2D Object Detection can be split into two categories: one-stage methods and two-stage methods. The two-stage methods (\!\!\cite{he2018mask, ren2016faster, cai2019cascade}) follow a proposal-driven approach: in the first stage a set of region proposals is generated and in the second stage the candidate locations are classified as objects or background by using a convolutional neural network and are also refined. Even though the accuracy is great, they lack in terms of inference speed, fueling the need of one-stage object detectors. The one-stage methods (\!\!\cite{redmon2016look,yolov10,lin2018focal}) perform both object localization and classification in a single pass through the network by relying on predefined anchor boxes.

In the context of 3D Object Detection, the focus shifts from identifying objects solely in 2D space to capturing their full spatial extent and orientation within a 3D environment. Unlike 2D detection, where bounding boxes suffice, 3D detection also requires predicting the objects' orientation relative to the coordinate system, along with their 3D bounding boxes. Some common types of inputs used for neural networks designed to solve this task are: point clouds, voxel grids, depth maps, RGB-D images and multi-view images.

The most challenging way of predicting the 3D cuboids is by solving the Monocular 3D Object Detection task, which involves using a single image as input, lacking in depth cues, which renders monocular object-level depth estimation naturally ill-posed. The standard network pipeline used for solving this task consists of a convolutional feature extraction backbone followed by a detection module added on top of it for determining the relevant 3D attributes needed to describe the 3D cuboids that represent the detected objects (\!\!\!\cite{brazil2023omni3d, kumar2022deviant, lu2021geometry, multivar}). However, convolutional neural networks can sometimes struggle with capturing long-range dependencies and contextual information due to their limited receptive field. This limitation can impact the network's ability to fully understand the spatial relationships within the image, which is crucial for accurate 3D object detection. 

Recently, transformer-based models have shown promise in various vision tasks by effectively modeling long-range dependencies and capturing global context. Incorporating transformers for feature extraction could potentially enhance the performance of monocular 3D object detection systems by addressing the limitations of CNNs and providing a more comprehensive understanding of the scene.

This paper proposes an original pipeline for Monocular 3D Object Detection built upon the following key contributions:
\begin{itemize}
    \item We introduce and validate a novel 3D Object Detection method based on a Transformer architecture as a feature extraction backbone.
    \item We leverage the power of the attention mechanism in the context of Monocular 3D Object Detection.
    \item Comprehensive experimental validation demonstrates that the proposed method outperforms the existing techniques, in both accuracy and robustness.
\end{itemize}

\section{Related work}
\subsection{Vision Transformers}
Vision Transformer (ViT) \cite{dosovitskiy2021image} represents the starting point when it comes to designing Transformer architectures suitable for different computer vision tasks, being the first architecture to prove that a pure Transformer can be applied on image patches and perform well in image classification without relying on CNNs. The most notable architectures of this kind are Swin-Transformer (Swin-T) \cite{liu2021swin}, Pyramid Vision Transformer (PVT) (\!\!\!\cite{Wang_2022, wang2021pyramid}) and Vision Transformer Adapter (ViT-Adapter) \cite{chen2023vision}.

Swin Transformer \cite{liu2021swin}, further improved in \cite{liu2022swin}, is a hierarchical Transformer which uses shifted windows in order to compute the representation for the image patches. Swin-T is compatible with various tasks, suck as image classification, object detection and semantic segmentation and it outputs feature maps at different scales in a hierarchical approach. Pyramid Vision Transformer \cite{wang2021pyramid}, further improved in \cite{Wang_2022}, uses multiple Transformer encoders, generates multi-scale feature maps and can also be used for multiple vision tasks. It is divided into four stages, each one being composed of a patch embedding layer and a Transformer encoder layer, generating feature maps of different scales. The authors from \cite{chen2023vision} propose a dense prediction task pre-training-free adapter for ViT, used to introduce the necessary inductive biases into the model for solving specific vision tasks, including object detection, instance segmentation and semantic segmentation. They use a spatial prior module which collects spatial features of three target resolutions, flattens and concatenates them in order to be fed as input for feature interaction. 

\subsection{Monocular 3D Object Detection}

Generally, 3D Object Detection is studied for two different domains, depending on the input data: autonomous systems (\!\!\cite{kumar2022deviant, lu2021geometry, multivar}) or indoor scenes (\!\!\cite{tulsiani2018factoring, nie2020total3dunderstanding}). The assumptions made in the methods solving this task are directly correlated to the domain in which they are applied.

Due to the fact that monocular 3D object detection is an ill-posed problem, Lu et al. \cite{lu2021geometry} propose a solution for urban scenes that includes a Geometry Uncertainty Projection (GUP) module and a Hierarchical Task Learning (HTL) strategy in order to tackle the depth inference error problem and the training instability due to the dependency between tasks. The authors from \cite{kumar2022deviant} propose a Depth EquiVarIAnt Network (DEVIANT) which uses scale equivariant steerable (SES) blocks \cite{sosnovik2020scaleequivariant} for the first time in the context of Monocular 3D Object Detection in order to learn consistent depth estimates by producing 5D feature maps in which the extra dimension captures the changes in scale for depth. Both approaches use uncertainty modeling to predict the physical and visual heights, but they do not model a joint probability distribution between the two, consequently lacking in information about their correlation. The authors from \cite{multivar} propose learning a full covariance matrix during training, with the guide of a multivariate likelihood. A general-purpose baseline method is proposed in \cite{brazil2023omni3d}, called Cube R-CNN, designed for solving the 3D Object Detection task for both data domains: outdoor and indoor scenes, which surpasses prior best approaches on various datasets. 

While convolutional neural networks have been the backbone of many successful 3D object detection methods, their limitations have become more apparent in recent years, reaching a performance plateau in image-based 3D object detection. Specifically, CNNs often struggle with capturing long-range dependencies and contextual information due to their inherently local receptive fields  \cite{liu2021swin}, which can restrict their ability to understand complex scenes fully. Additionally, CNNs can be less efficient in handling varying scales and aspect ratios of objects, leading to potential performance degradation in diverse environments. As a result, there is a growing need to explore more advanced architectures, such as transformers, which can model global context and offer greater flexibility in feature extraction.

To this end, some authors experiment with Transformer-based architectures (\!\!\!\cite{zhang2023monodetr, he2023ssdmonodetr, he2023s3monodetr}). The authors of \cite{zhang2023monodetr} design MonoDETR which uses a depth-aware transformer that guides the detection process by integrating contextual depth cues along with the visual features obtained from the input image. Based on the depth-aware transformer, the authors from \cite{he2023ssdmonodetr} propose a new attention mechanism for Monocular 3D Object Detection called Supervised Scale-aware Deformable Attention (SSDA) which uses preset masks with different scales and a Weighted Scale Matching (WSM) loss to supervise scale prediction. However, the reliance on contextual depth cues might limit its performance in scenarios with insufficient or misleading depth information. As an improvement, in \cite{he2023s3monodetr}, a more complex Supervised Shape\&Scale-perceptive Deformable Attention ($S^3$-DA) and a Multi-classification-based Shape\&Scale Matching (MSM) loss are proposed, which extract features of different shapes and scales. This method, while advanced, may introduce additional computational overhead and complexity. 
\begin{figure*}
\includegraphics[width=\textwidth,height=7cm]{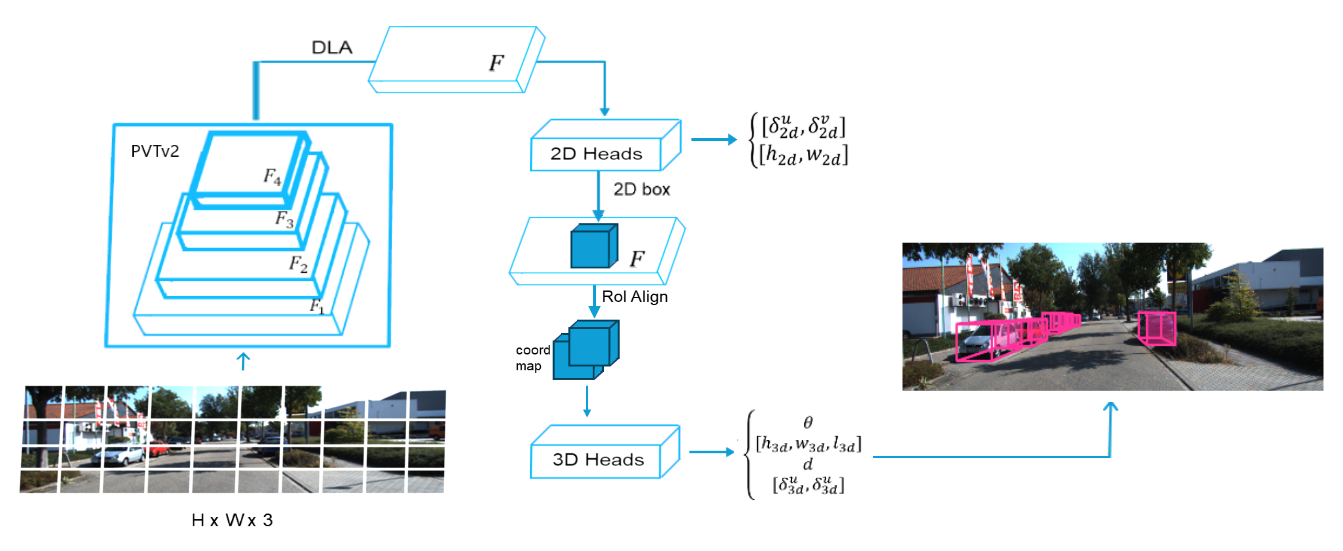}
  \caption{Architecture overview. LAM3D takes as input a single image, extracts feature maps $\{F_1, F_2, F_3, F_4\}$ of different scales with a Transformer-based backbone and aggregates them using DLA into a single feature map $F$ further used to infer the 2D and 3D bounding box parameters. }
  \label{fig:network_pipeline}
\end{figure*}

It is important to remark that, even though all previously mentioned methods utilize the Transformer architecture, they use it exclusively for guiding the detection part of the pipeline and not as a feature extractor backbone. To the best of our knowledge, the authors from \cite{dst3d} are the only ones that use a Transformer-based architecture instead of a CNN for extracting the features from the input image. They propose DST3D, based on DLA-Swin Transformer (DST) as a feature extractor backbone and trained end-to-end. The DLA-Swin Transformer is based on Swin-T \cite{liu2021swin} and uses Deep Layer Aggregation (DLA) \cite{yu2019deep} for better feature fusion between different layers and blocks. However, the Pyramid Vision Transformer (PVT) offers several advantages over DLA-Swin, such as a more efficient multi-scale feature extraction (\!\!\!\cite{wang2021pyramid}). PVT's hierarchical design and attention mechanisms provide a more flexible and powerful approach to feature extraction, making it a superior choice for monocular 3D object detection.

\section{Method}
Our goal is to design an innovative and effective method for Monocular 3D Object Detection in the context of real-world traffic scenery, which benefits from using the attention mechanism in extracting features from the input. Our approach extends PVTv2 \cite{Wang_2022} by incorporating detection heads and loss functions that have been validated for this task. We refer to our method as LAM3D. Figure \ref{fig:network_pipeline} shows an overview of our approach.

The input image is firstly processed by a Transformer-based 2D detection backbone. The resulting 2D bounding boxes as Regions of Interest (RoIs) are further used as input for the convolutional 3D heads which extract the 3D bounding box information in terms of size, angle and 3D projected center. The depth is inferred afterwards by using a Geometry Uncertainty Projection module, based on the 2D and 3D heights and a depth bias, as previously described in \cite{lu2021geometry}.

This section presents the detection pipeline, starting with the Transformer-based feature extraction backbone.

\subsection{Transformer Backbone}
The Pyramid Vision Transformer architecture \cite{wang2021pyramid}, similar to convolutional neural network backbones, outputs a feature pyramid represented by four feature maps of different scales $\{F_1, F_2, F_3, F_4\}$, suitable for dense prediction tasks. The resulted feature pyramid is a consequence of the four stages of the Pyramid Vision Transformer, all of them sharing a similar architecture formed of a patch embedding layer and $L_i$ Transformer encoder layers. At the first stage, the input image is divided into flattened patches which are projected into a $C_1$-dimensional embedding. The embedded patches together with a position embedding are fed to the Transformer encoder layers and the output is represented by a feature map of size $\frac{H}{4} \times \frac{W}{4} \times C_1$. The process is repeated at each stage with the previously generated feature map as input. Following the same concept, each feature map has the shape $\frac{H}{s_i} \times \frac{W}{s_i} \times C_i$, where $s$ is the stride at stage $i$, with respect to the input image, and $C_i$ is the embedding dimension at stage $i$. The strides used at each stage in order to generate the feature pyramid are $\{4,8,16,32\}$ pixels.

The Pyramid Vision Transformer was previously validated as backbone for the 2D Object Detection task with COCO 2017 as primary benchmark \cite{wang2021pyramid}. In the context of this paper, the Transformer backbone is initialized with the weights pre-trained on ImageNet \cite{imagenet} and fine-tuned on KITTI 3D Object Detection Dataset for the 2D Object Detection sub-task for baselines 1 and 2.

\subsection{Neck}
The neck is represented by a fully convolutional upsampling method which uses iterative deep aggregation (IDA) to refine resolution and aggregate scale stage-by-stage \cite{yu2019deep}. The feature map is then upsampled using bilinear interpolation. Throughout our experiments, we observed that the first 64 channels of the feature map were the most critical. By slicing and focusing on these essential channels, we can reduce computational complexity and memory usage while retaining the most significant information for the detection task.

\subsection{Detection Heads}
The proposed network uses three 2D detection heads for determining the heatmap, offset and size for each potential 2D bounding box. The heatmap represents the coarse locations and the level of confidence given to each region in the image in terms of each class, with shape [W,H,C] where W, H are the width, respectively the height of the input image and C is the number of classes. The coarse locations for each object, computed based on the generated heatmap, are refined to the bounding box center by a 2D offset head which computes a refinement bias $(\delta_{2d}^u, \delta_{2d}^v)$. The size for each 2D bounding box $(w_{2d}, h_{2d})$ is computed by a 2D size head. 

The features corresponding to the Region of Interest (RoI) are cropped afterwards so only information from the object level is kept in the next processing steps. The 3D Detection convolutional heads are used to infer the 3D bounding box information: rotation angle, 3D dimensions, and 3D center projection. The 3D offset head determines the 3D center projection $(\delta_{3d}^u, \delta_{3d}^v)$, the angle head determines the $\theta$ rotation angle and the 3D size head determines the 3D bounding box parameters $(h_{3d}, w_{3d}, l_{3d})$, where $h_{3d}$ is the height, $w_{3d}$ is the width and $l_{3d}$ is the length. 

In order for the 3D bounding box to be completely defined, the depth also has to be inferred. Given that regressing the depth directly is difficult and prone to errors, a Geometry Uncertainty Projection module \cite{lu2021geometry} is used for inferring the depth in the context of a well-defined probability framework.

\subsection{Loss Functions}
The total loss of the model $L_{total}$ can be defined as a sum of all the task losses as per Eq. \ref{eq:total_loss}.
\begin{equation}
\begin{aligned}
    L_{total} = L_{heatmap} + L_{offset2d} + L_{size2d} + L_{angle} \\ + L_{w3d} + L_{l3d} + L_{h3d} + L_{depth} + L_{offset3d}
\end{aligned}
\label{eq:total_loss}
\end{equation}

L1 Loss Function is used as $L_{offset2d}$, $L_{size2d}$, $L_{offset3d}$. In the context of $L_{size3d}$, L1 Loss Function is used only for $w_{3d}$ and $l_{3d}$ parameters, as the estimated height $h_{3d}$ is used in the Geometry Uncertainty Projection module. Cross-Entropy Loss Function is used as a term in $L_{angle}$ for guiding the object classification, together with the L1 Loss Function for angle value regression. $L_{heatmap}$ is equivalent to the focal loss \cite{lin2018focal}. In order to tackle the problem of training instability, each task should start training only after its pre-tasks have been well-trained.

\section{Experiments}
\subsection{Setup}
\subsubsection{Dataset}
KITTI 3D Object Detection \cite{KITTI} consists of 7481 training images and 7518 test images, as well as their corresponding point clouds, comprising a total of 80.256 labeled objects. Additionally to labeling the objects, each bounding box is marked as either visible, semi-occluded, fully occluded or truncated. The image resolution is 375 x 1242. A common approach, also followed in this paper, is to split the training data into a training set of 3712 images and a validation set of 3769 images \cite{lu2021geometry}. The following studies are conducted based on this split and the results shown in this chapter are on the validation subset.
\subsubsection{Evaluation protocol}
All the conducted experiments follow the evaluation configurations described in the KITTI 3D Object Detection Benchmark, using the PASCAL criteria. In the official benchmark evaluation, the 3D bounding box overlap for car category is 70\%, while for pedestrians and cyclists is 50\%. As an additional evaluation configuration, experiments in which the 3D bounding box overlap for car category is 50\% and for pedestrians and cyclists is 30\% were also conducted.
\subsubsection{Implementation details}
The resolution of the input image for the proposed method is 380 x 1280. The proposed model is trained for 140 epochs with a batch size of 12 on 2 NVIDIA GeForce RTX 4090 GPUs. The training optimizer is Adam with an initial learning rate of $1.25 \times 10^{-3}$, which is decayed at epochs \{90, 120\} with a rate set to 0.1. In the first 5 epochs, a cosine warm-up is applied. The Transformer backbone pre-trained on ImageNet \cite{imagenet} is firstly fine-tuned on KITTI 3D Object Detection Dataset for the 2D Object Detection sub-task. The experiments were made with PVTv2 baselines 1 and 2 \cite{Wang_2022}.

\begin{figure*}[htbp!]
    \centering
    \subfloat[LAM3D (Ours)]{\includegraphics[width=0.45\textwidth]{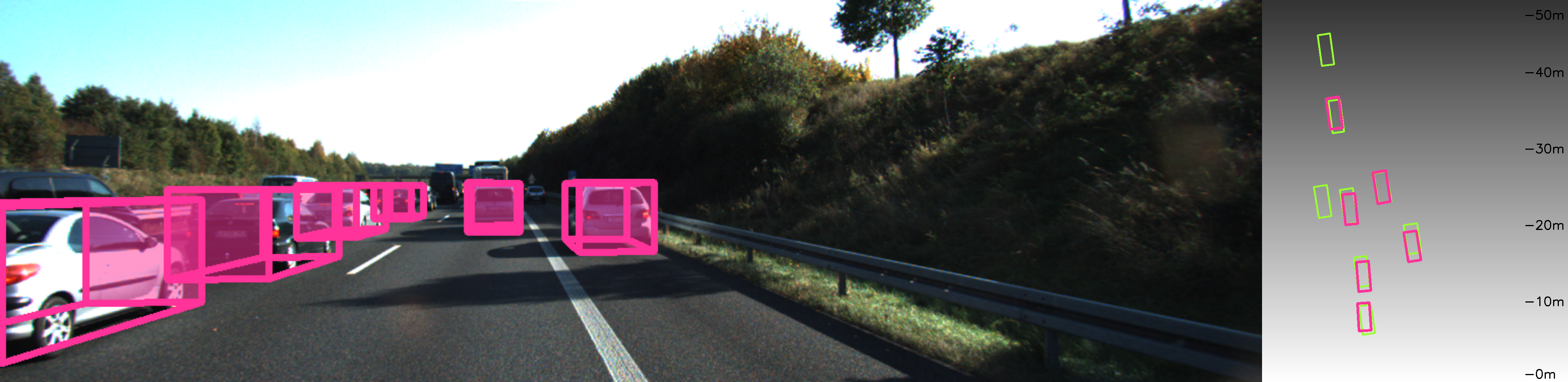}\label{fig:1054}}
    \hspace{0.02\textwidth}
    \subfloat[GUP Net* \cite{lu2021geometry}]{\includegraphics[width=0.45\textwidth]{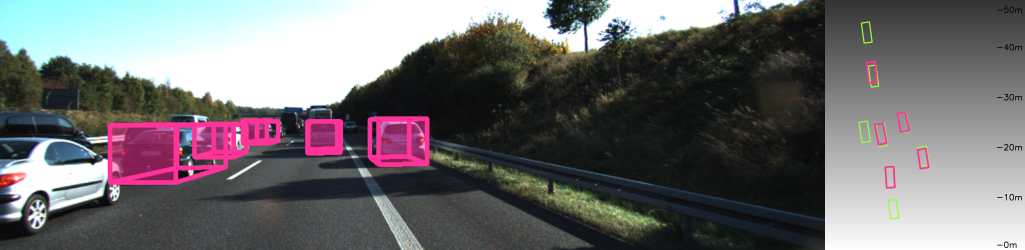}\label{fig:1054_b}} \\
    \subfloat[LAM3D (Ours)]{\includegraphics[width=0.45\textwidth]{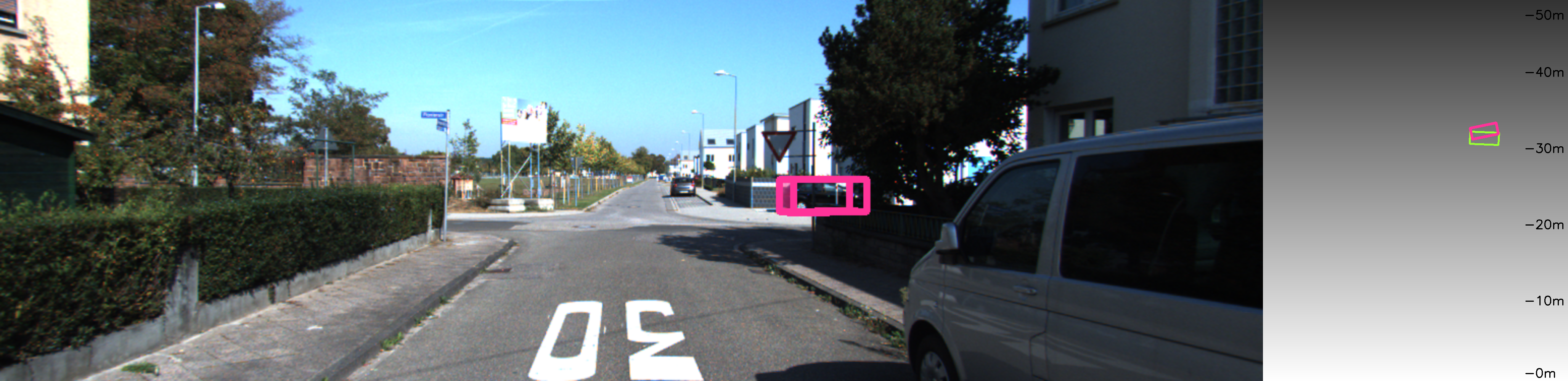}\label{fig:3456}}
    \hspace{0.02\textwidth}
    \subfloat[GUP Net* \cite{lu2021geometry}]{\includegraphics[width=0.45\textwidth]{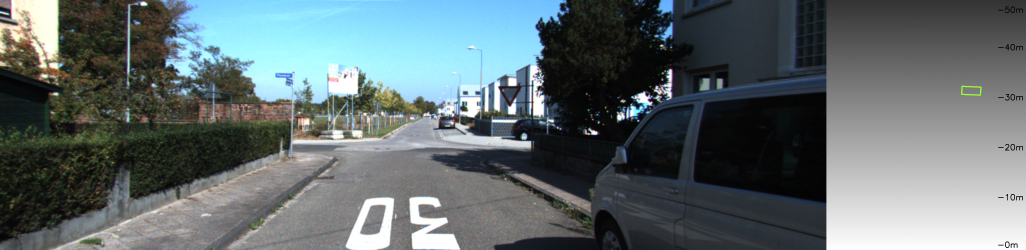}\label{fig:3456_b}}
    \caption{Qualitative results for truncated and occluded objects. The ground truth is represented by green bounding boxes and the pink bounding boxes represent the predictions of different methods.}
    \label{fig:qualitative_out}
\end{figure*}

\subsection{Results}
\label{cap:main_results}

We report the evaluation results of our method on the KITTI Dataset in Tab. \ref{table:3d_dst3d} against the best Transformer-based method evaluated on KITTI, namely DST3D \cite{dst3d}. As commonly done, we report the results on the official test split where the evaluation is done by KITTI servers using withheld ground truth. Tab. \ref{table:scale_aug_ap} reports the results on the validation set for further evaluation. As can be seen, our proposed method achieves superior performance than previous Transformer-based methods for all difficulty levels. Specifically, LAM3D gains significant improvements in both $AP_{3D}$  and $BEV_{3D}$. Tab. \ref{table:3d_dst3d} shows that the average precision @70 produced by LAM3D improves to 19.85\% for Easy, 12.86\% for Moderate, and 10.94\% for Hard scenarios, compared to 7.02\%, 3.58\%, and 3.58\% over  DST3D respectively. The same behavior is observed for AP@50. This demonstrates the superior capability of Pyramidal Transformers-based feature extractors compared to Swin-T on Object Detection tasks for large scenes.

\begin{table}[ht!]
\caption{Comparison of Transformers-based Architectures on KITTI Test cars}

\centering
\resizebox{\columnwidth}{!}{%
\begin{tabular}{@{}lllllccccccc@{}}
\toprule
\multirow{2}{*}{Method}  & \multirow{2}{*}{Backbone} & \multicolumn{3}{c}{$AP_{3D}(IOU=0.7|R_{40})$} & \multicolumn{3}{c}{$AP_{3D}(IOU=0.5|R_{40})$} \\ \cmidrule(lr){3-5} \cmidrule(lr){6-8}
 &  &  Easy & Mod. & Hard & Easy & Mod. & Hard \\ \midrule
DST3D \cite{dst3d}  & DLA-Swin-T & 7.02 & 3.58 & 2.96 & 12.73 & 8.23 & 6.68 \\ \midrule
\textbf{LAM3D (Ours)}   & PVTv2 b2 & \textbf{19.87} & \textbf{12.86} & \textbf{10.94} & \textbf{28.49} & \textbf{18.92} & \textbf{15.80} \\ \bottomrule
\end{tabular}%
}
\label{table:3d_dst3d}              
\end{table}
In Tab. \ref{table:scale_aug_ap} and  \ref{table:scale_aug_var}, we compare our method to GUP Net \cite{lu2021geometry}, which also employs the scale augmentation principle through the Geometry Uncertainty Projection model. Our results demonstrate that for the same image resolution, the features extracted by PVT provide superior benefits for pose estimation. Tab. \ref{table:scale_aug_ap} highlights the performance of both methods on the KITTI validation set for cars, pedestrians, and cyclists. Our method, LAM3D, demonstrates improved Average Precision (AP) for both 3D and Bird's Eye View (BEV) metrics across various difficulty levels (Easy, Moderate, Hard). Notably, LAM3D outperforms GUP Net in the car category with higher AP scores in most scenarios with an improvement of 1.48\% at $AP_{70}$ and 1.15\% at $AP_{50}$. Additionally, LAM3D exhibits competitive results for pedestrians and cyclists, achieving better AP scores in the majority of cases. Tab. \ref{table:scale_aug_var} reports the variance in 3D predictions for the car category. Our method achieves lower variance values compared to GUP Net, indicating more consistent and reliable predictions. This consistency is evident in both the 3D and BEV metrics across all difficulty levels, further establishing the robustness of our approach.

\begin{table}[ht!]
\caption{Comparison of equivalent Architectures on KITTI Val. The methods denoted with * were retrained.}
\centering
\resizebox{\columnwidth}{!}{%
\begin{tabular}{@{}lllllccccccc@{}}
\toprule
\multirow{2}{*}{Method}  & \multirow{2}{*}{Category} & \multicolumn{3}{c}{$AP_{3D}(IOU=0.7|R_{40})$} & \multicolumn{3}{c}{$AP_{3D}(IOU=0.5|R_{40})$} \\ \cmidrule(lr){3-5} \cmidrule(lr){6-8}
 &  &  Easy & Mod. & Hard & Easy & Mod. & Hard \\ \midrule
 & Car & {21.07}  & {15.44} & \textbf{12.88} & {58.97} & {44.00} & {38.08} \\ 
GUP Net* \cite{lu2021geometry}  & Pedestrian & \textbf{{9.27}}  & \textbf{{6.84}} & \textbf{5.73} & {25.85} & {20.66} & {16.99} \\ 
  & Cyclist & {4.44}  & {2.18} & {2.03} & {18.06} & {10.19} & {9.08} \\ \midrule

  & Car & \textbf{22.55} & \textbf{15.66} & {12.85} & \textbf{60.08} & \textbf{45.62} & \textbf{39.50} \\ 
{LAM3D (Ours)}   & Pedestrian & {8.98} & {6.64} & {5.33} & \textbf{26.44} & \textbf{20.81} & \textbf{17.07} \\ 
   & Cyclist & \textbf{8.09} & \textbf{3.87} & \textbf{3.78} & \textbf{21.48} & \textbf{11.19} & \textbf{11.01} \\ \bottomrule
   
\end{tabular}%
}
\label{table:scale_aug_ap}      
\end{table}

\begin{table}[ht!]
\caption{Variance values of the 3D predictions for the \textit{car} category. The methods denoted with * were retrained.}
\centering
\resizebox{\columnwidth}{!}{%
\begin{tabular}{@{}lllllccccccc@{}}
\toprule
\multirow{2}{*}{Method} & \multirow{2}{*}{Input}   & \multicolumn{3}{c}{$\sigma_{3D}(IOU=0.7|R_{40})$} & \multicolumn{3}{c}{$\sigma_{3D}(IOU=0.5|R_{40})$} \\ \cmidrule(lr){3-5} \cmidrule(lr){6-8}
 &  &  Easy & Mod. & Hard & Easy & Mod. & Hard \\ \midrule
GUP Net* \cite{lu2021geometry} & Image & 7.77  & 6.47 & 5.55 & 16.08 & 17.10 & 16.86 \\
Ours & Image & \textbf{5.73 }  & \textbf{5.60} & \textbf{4.86} & \textbf{14.27} & \textbf{16.44} & \textbf{16.06} \\ \bottomrule
\end{tabular}%
}
\label{table:scale_aug_var}            
\end{table}

\subsection{Ablation Study}
In order to validate the effectiveness of using a Transformer-based architecture instead of a Convolutional Neural Network backbone in the context of the reference method used, ablation studies detailed in this section were performed on the validation split for KITTI 3D Object Detection Benchmark.

\subsubsection{Attention mechanism}
The results can be seen in Tab. \ref{table:att_influence_2d} for the 3D Object Detection task with PVTv2 backbone. This table proves the efficiency of the attention mechanism in the Transformer-based architecture.
\begin{table}[ht!]
\caption{Evaluation of Attention's impact on KITTI Val.}
\centering
\resizebox{\columnwidth}{!}{%
\begin{tabular}{@{}cllllccccccc@{}}
\toprule
\multirow{2}{*}{Attention}  & \multirow{2}{*}{Category} & \multicolumn{3}{c}{$AP_{3D}(IOU=0.7|R_{40})$} & \multicolumn{3}{c}{$AP_{3D}(IOU=0.5|R_{40})$} \\ \cmidrule(lr){3-5} \cmidrule(lr){6-8}
 &  &  Easy & Mod. & Hard & Easy & Mod. & Hard \\ \midrule
& Car & 18.61 & 13.28 & 10.87 & 57.10 & 42.87 & 36.97 \\
\ding{55} & Pedestrian & 7.33 & 5.54 & 4.46 & 24.44 & 19.45 & 15.61 \\
& Cyclist & 4.32 & 2.55 & 2.10 & 14.79 & 8.20 & 7.02  \\ \hline
  & Car & \textbf{22.55} & \textbf{15.66} & \textbf{{12.85}} & \textbf{60.08} & \textbf{45.62} & \textbf{39.50} \\ 
\ding{51}   & Pedestrian & \textbf{{8.98}} &\textbf{ {6.64}} & \textbf{{5.33}} & \textbf{26.44} & \textbf{20.81} & \textbf{17.07} \\ 
   & Cyclist & \textbf{8.09} & \textbf{3.87} & \textbf{3.78} & \textbf{21.48} & \textbf{11.19} & \textbf{11.01} \\ \bottomrule 
\end{tabular}%
}
\label{table:att_influence_2d}              
\end{table}

\section{Qualitative results}
In Figure \ref{fig:qualitative_out}, we introduce two visualizations for truncated and occluded vehicles, highlighting the challenges of autonomous driving scenarios. One can note that our method effectively addresses these challenges, demonstrating its robust performance compared to GUP Net \cite{lu2021geometry}.

\section{Conclusion}
In this paper, we propose LAM3D, a novel framework for solving the Monocular 3D Object Detection task. The results detailed in Section \ref{cap:main_results} prove that Transformers can be confidently used as feature extraction backbones in the context of the ill-posed Monocular 3D Object Detection task, achieving similar or even better performance than CNNs. Furthermore, the ablation study conducted indicates that the attention mechanism is particularly effective at capturing relevant information, especially for smaller objects in the scene.

\section*{Acknowledgment}
This work is funded by Innoviris within the research project TORRES.\\

\bibliographystyle{ieeetr}
\bibliography{ref.bib}

\end{document}